\documentclass[runningheads]{llncs}
\usepackage{graphicx}

\usepackage{tikz}
\usepackage{comment} 
\usepackage{amsmath,amssymb} 
\usepackage{color}
\usepackage{hyperref}
\usepackage{gensymb}
\usepackage[ruled, vlined]{algorithm2e} 
\usepackage[abs]{overpic}
\usepackage{url}
\usepackage{cite}

\newcommand\tabcaption{\def\@captype{table}\caption} 
\newcommand{\etal}{et al. }
\newcommand{\para}[1]{\noindent\textbf{{#1}}~~}

\begin{document}
\pagestyle{headings}
\mainmatter
\def\ECCVSubNumber{1163}  

\title{Modeling 3D Shapes by Reinforcement Learning} 

\titlerunning{Modeling 3D Shapes by Reinforcement Learning}

%
\author{Cheng Lin\inst{1,2}\and
Tingxiang Fan\inst{1}\and
Wenping Wang\inst{1}\and
Matthias Nie\ss ner\inst{2}}
\authorrunning{C. Lin et al.}
%
\institute{The University of Hong Kong \and Technical University of Munich 
 \\
}
\maketitle

\begin{abstract}
We explore how to enable machines to model 3D shapes like human modelers using deep reinforcement learning (RL). In 3D modeling software like Maya, a modeler usually creates a mesh model in two steps: (1) approximating the shape using a set of primitives; (2) editing the meshes of the primitives to create detailed geometry. Inspired by such artist-based modeling, we propose a two-step neural framework based on RL to learn 3D modeling policies.  By taking actions and collecting rewards in an interactive environment, the agents first learn to parse a target shape into primitives and then to edit the geometry. To effectively train the modeling agents, we introduce a novel training algorithm that combines heuristic policy, imitation learning and reinforcement learning.  Our experiments show that the agents can learn good policies to produce regular and structure-aware mesh models, which demonstrates the feasibility and effectiveness of the proposed RL framework.
\end{abstract}

\section{Introduction}

Enabling machines to learn the behavior of humans in visual arts, such as teaching machines to paint \cite{ganin2018spiral, riaz2018learningdeepsketchRL, huang2019learning}, has aroused researchers' curiosity in recent years. The 3D modeling, a process of preparing geometric data of 3D objects, is also an important form of visual and plastic arts and has wide applications in computer vision and computer graphics. Human modelers are able to form high-level interpretations of 3D objects, and use them for communicating, building memories, reasoning and taking actions. Therefore, for the purpose of enabling machines to understand 3D artists' behavior and developing a modeling-assistant tool, it is a meaningful but under-explored problem to teach intelligent agents to learn 3D modeling policies like human modelers.

\begin{figure} 
    \centering      
  \begin{overpic}[width=0.93\linewidth]{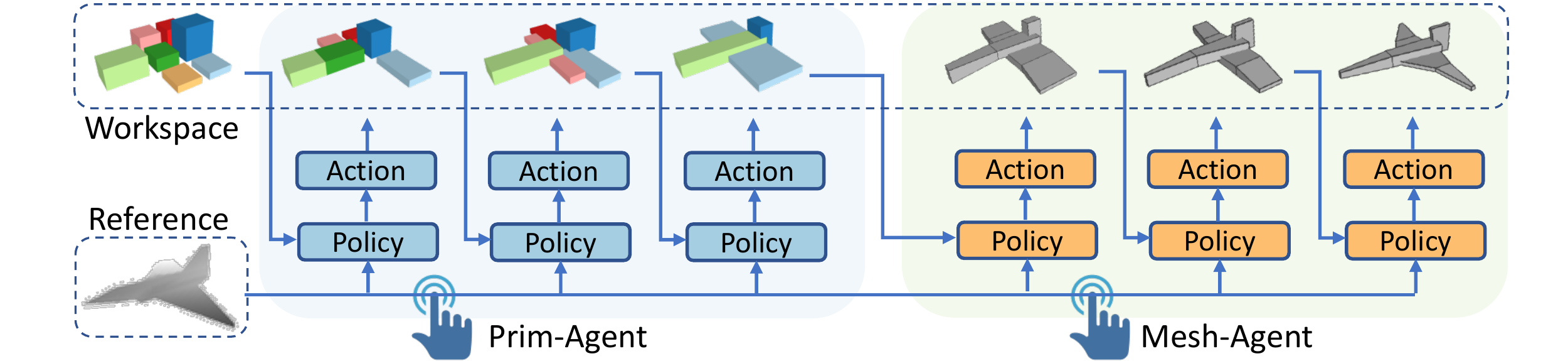}
    \end{overpic}
  \caption{The RL agents learn policies and take actions to model 3D shapes like human modelers. Given a reference, the Prim-Agent first approximates the shape using primitives, and then the Mesh-Agent edits the mesh to create detailed geometry.}
  \label{fig:teaser}
\end{figure}

Generally, there are two steps for a 3D modeler to model a 3D shape in mainstream modeling software. First, the modeler needs to perceive the part-based structure of the shape, and starts with basic geometric primitives to approximate the shape. Second, the modeler edits the mesh of the primitives using specific operations to create more detailed geometry. These two steps embody humans' hierarchical understanding and preserve high-level regularity within a 3D shape, which is more accessible compared to predicting low-level points. 

Inspired by such artist-based modeling, we propose a two-step deep reinforcement learning (RL) framework to learn 3D modeling policies. RL is a decision-making framework, where an agent interacts with the environment by executing actions and collecting rewards. As visualized in Fig.~\ref{fig:teaser}, in the first step, we propose Prim-Agent which learns to parse a target shape into a set of primitives. In the second step, we propose Mesh-Agent to edit the meshes of the primitives to create more detailed geometry.

There are two major challenges to teach RL agents to model 3D shapes. The first one is the environment setting of RL for shape analysis and geometry editing. The Prim-Agent is expected to understand the shape structure and decompose the shape into components. For this task, however, the interaction between agent and environment is not intuitive and naturally derived. To motivate the agent to learn, rather than directly predicting the primitives, we break down the main task into small steps; that is, we make the agent operate a set of pre-defined primitives step-by-step to approximate the target shape, which finally results in a primitive-based shape representation. For the Mesh-Agent, the challenge lies in preserving mesh regularity when editing geometry. Instead of editing single vertices, we propose to operate the mesh based on edge loops \cite{edgeloop2000}. Edge loop is a widely used technique in 3D modeling software to manage complexity, by which we can edit a group of vertices and control an integral geometric unit. The proposed settings capture the insights of the behavior of modeling artists and are also tailored to the properties of the RL problem.

The second challenge is, due to the complex operations and huge action space in this problem, off-the-shelf RL frameworks are unable to learn good policies. Gathering demonstration data from human experts to guide the agents would help, but this modeling data is expensive to obtain, while the demonstrations are far from covering most scenarios the agent will experience in real-world 3D modeling. To address this challenge, innovations are made on the following two points. First, we design a heuristic algorithm as a ``virtual expert'' to generate demonstrations, and show how to interactively incorporate the heuristics into an imitation learning (IL) process. Second, we introduce a novel scheme to effectively combine IL and RL for modeling 3D shapes. The agents are first trained by IL to learn an initial policy, and then they learn in an RL paradigm by collecting the rewards. We show that the combination of IL and RL gives better performance than either does on its own, and it also outperforms the existing related algorithms on the 3D modeling task.

To demonstrate our method, we condition the modeling agents mainly on the shape references from single depth maps. Note, however, the architecture of our agents is agnostic to the shape reference, while we also test RGB images. The contributions of this paper are three-fold: 

\begin{itemize}
    \item We make the first attempt to study how to teach machines to model real 3D shapes like humans using deep RL. The agents can learn good modeling policies by interacting with the environment and collecting feedback.
    \item We introduce a two-step RL formulation for shape analysis and geometry editing. Our agents can produce regular and structure-aware mesh models to capture the fundamental geometry of 3D shapes. 
    \item We present a novel algorithm that combines heuristic policy, imitation learning and reinforcement learning. We show a considerable improvement compared to the related training algorithms on the 3D modeling task. 
\end{itemize}

\section{Related Work}
\para{Imitation learning and reinforcement learning} Imitation learning (IL) aims to mimic human behavior by learning from demonstrations. Classical approaches \cite{abbeel2004apprenticeship, ziebart2008maximum} are based on training a classifier or regressor to predict behavior using demonstrations collected from experts. However, since policies learned in this way can easily fail in theory and practice \cite{ross2010efficient}, some interactive strategies for IL are introduced such as DAagger \cite{ross2011dagger} and AggreVaTe \cite{ross2014reinforcement}.

Reinforcement learning (RL) is to train an agent by making it explore in an environment and collect rewards. With the development of the scalability of deep learning \cite{lecun2015deep}, a breakthrough of deep reinforcement learning (DRL) is made by the introduction of Deep Q-learning (DQN) \cite{mnih2015natureRL}. Afterward, a series of approaches have been continuously proposed to improve the DQN, such as Dueling DQN \cite{duedlingdqn}, Double DQN \cite{van2016doubledqn} and Prioritized experience replay \cite{schaul2015PER}. 

Typically, an RL agent can find a reasonable action only after numerous steps of poor performance in exploration, which leads to low learning efficiency and accuracy. Thus, there has been interest in combining IL with RL to achieve better performance \cite{cruz2017pretrain1, subramanian2016pretrain2, silver2016mastering}. For example, Hester \etal proposed Deep Q-learning from Demonstrations (DQfD) \cite{dqfd2018}, in which they initially pre-train the networks solely on the demonstration data to accelerate the RL process. However, our experiments show that directly using these approaches for 3D modeling does not produce good performance; thus we introduce a novel variant algorithm that enables the modeling agents to learn considerably better policies.

\para{Shape generation by RL} Painting is an important form for people to create shapes. There is a series of methods using RL to learn how to paint by generating strokes \cite{xie2013strokeink, ganin2018spiral, huang2019learning} or drawing sketches \cite{zhou2018learningtodoodle, riaz2018learningdeepsketchRL}. 
Some works explore to use grammar parsing for shape analysis and modeling. Teboul \etal \cite{teboul2011shapegrammarparsing} use RL to parse the shape grammar of the building facade. Ruiz-Montiel \etal \cite{ruiz2013design} propose an approach to complement the generative power of shape grammars with RL techniques. These methods all focus on the 2D domain, while our method targets 3D shape modeling, which is under-explored and more challenging.  Sharma \etal \cite{sharma2018csgnet} present CSG-Net, which is a neural architecture to parse a 2D or 3D input into a collection of modeling primitives with operations. However, it only handles synthetic 3D shapes composed of the most basic geometries, while our method is evaluated on ShapeNet \cite{chang2015shapenet} models.

\begin{figure*} [!t]
    \centering
  \begin{overpic}[width=0.9\linewidth]{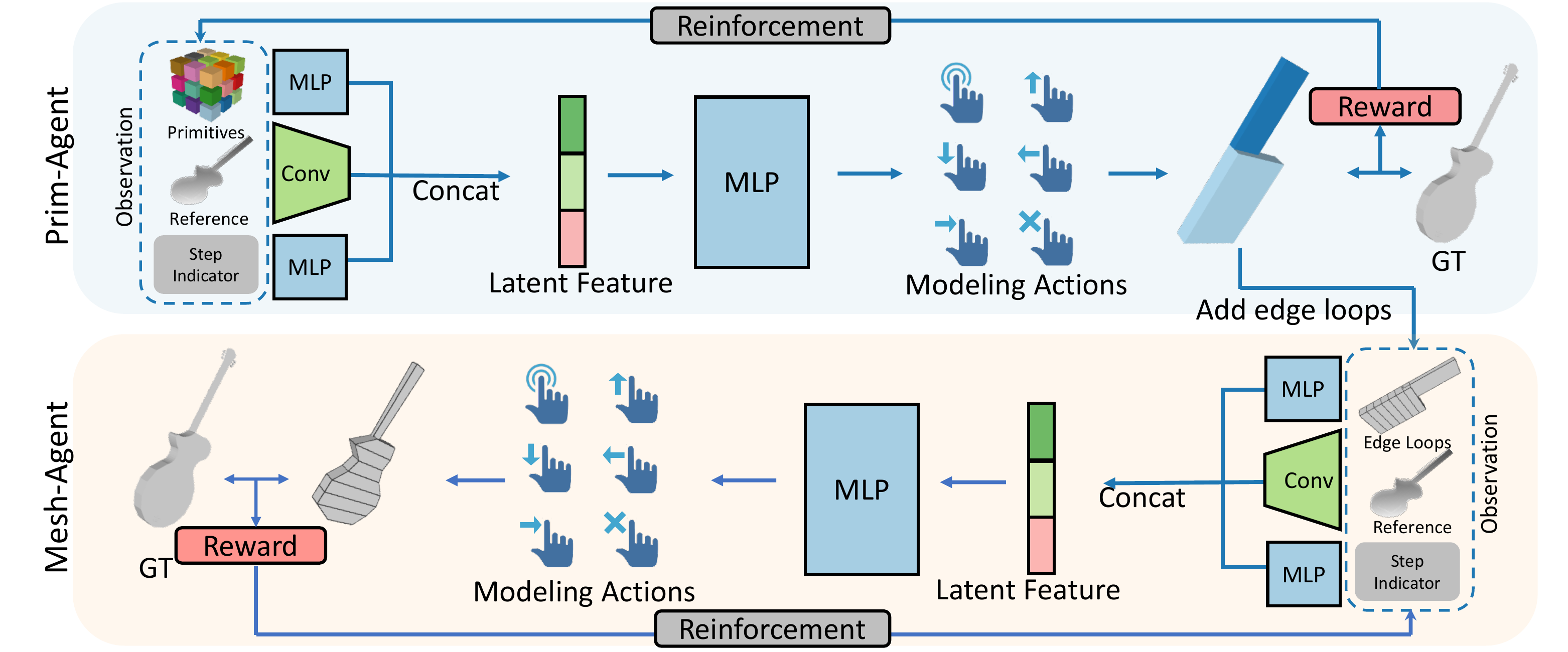}
    \end{overpic}
  \caption{The architecture of our two-step pipeline for 3D shape modeling. First, given a shape reference and pre-defined primitives, the Prim-Agent predicts a sequence of actions to operate the primitives to approximate the target shape. Then the edge loops are added to the output primitives. Second, the Mesh-Agent takes as input the shape reference and the primitive-based representation, and predicts actions to edit the meshes to create detailed geometry. }
  \label{fig:pipline}
\end{figure*} 

\para{High-level shape understanding} There has been growing interest in high-level shape analysis, where the ideas are central to part-based segmentation \cite{kalogerakis2010learning, Kalogerakis:2017:ShapePFCN} and structure-based shape understanding \cite{wang2011symmetrykaixu, li2017grass}. Primitive-based shape abstraction \cite{wu2005primitive, tulsiani2017learningboxprimitive, 3dprnn, paschalidou2019superquadrics}, in particular, is well-researched for producing structurally simple representation and reconstruction. Zou \etal \cite{3dprnn} introduce a supervised method that uses a generative RNN to predict a set of primitives step-by-step to synthesize a target shape. Li \etal \cite{li2017grass} and Sun \etal \cite{sun2019abstraction} propose neural architectures to infer the symmetry hierarchy of a 3D shape. Tian \etal \cite{tian2018execute3dprogram} propose a neural program generator to represent 3D shapes as 3D programs, which can reflect shape regularity such as symmetry and repetition. These methods capture higher-level shape priors but barely consider geometric details. Instead, our method performs joint primitive-based shape understanding and mesh detail editing. In essence, these methods have different goals with our work. They aim to directly minimize the reconstruction loss using end-to-end networks, while we focus on enabling machines to understand the environment, learn policies and take actions like human modelers.

\section{Method}
In this section, we first give the detailed RL formulations of the Prim-Agent (Sec.~\ref{sec:prim-agent}) and the Mesh-Agent (Sec.~\ref{sec:mesh_agent}). Then, we introduce an algorithm to efficiently train the agents (Sec.~\ref{sec:virtualexpert} and \ref{sec:training_strategy}). We will discuss and evaluate these designs in the next section.

\subsection{Primitive-based Shape Abstraction}
\label{sec:prim-agent}
The Prim-Agent is expected to understand the part-based structure of a shape by interacting with the environment. We propose to decompose the task into small steps, where the agent constantly tweaks the primitives based on the feedback to achieve the goal. The detailed formulation of the Prim-Agent is given below.

\begin{figure}[!htb]
\centering
\begin{overpic}[width=0.85\linewidth]{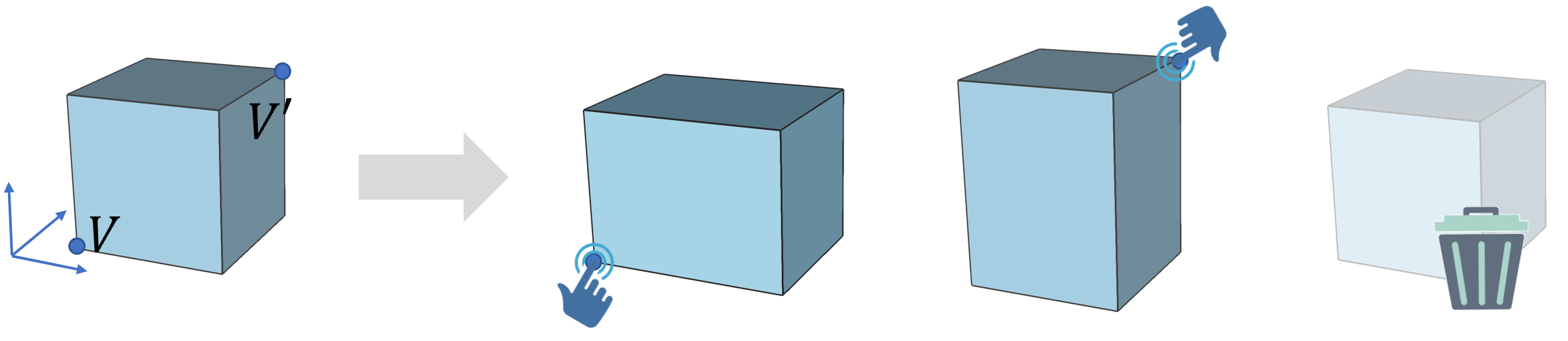}
        \put(10, -5) {Primitive $P_i$}
        \put(100, -5) {Edit corner $V$}
        \put(170, -5) {Edit corner $V'$}
        \put(250, -5) {Delete $P_i$}
\end{overpic}
\caption{Visualization of the three types of actions to operate a primitive.}
\label{fig:actions1}
\end{figure}
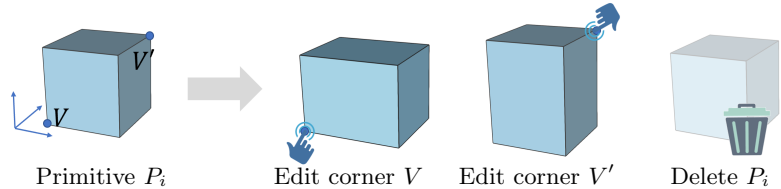 

\para{State} At the beginning, we arrange $m^3$ cubes that are uniformly distributed in the canonical frame ($m$ cubes for each axis), denoted as $\mathcal{P}=\{P_i~|~i=1,...,m^3\}$. We use $m=3$ in this paper. Each cuboid is defined by a six-tuple $(x,y,z,x',y',z')$ which specifies its two diagonal corner points $V=(x,y,z)$ and $V'=(x',y',z')$ (see Fig.~\ref{fig:actions1}). We define the state by: (1) the input shape reference; (2) the updated cuboid primitives at each iteration; (3) the step number represented by one-hot encoding.  The agent will learn to predict the next action by observing the current state.

\para{Action} As shown in Fig. \ref{fig:actions1}, we define three types of actions to operate a cuboid primitive $P_i$: (1) drag the corner $V$; (2) drag the corner $V'$; (3) delete $P_i$. For each type of action, we use four parameters $-2,-1,1,2$ to control the range of movement on the axis directions (for the delete action, these parameters all lead to deleting the primitive). In total, there are 27 cuboids, 3 types of actions, 3 moving directions ($x,y$ and $z$) for the drag actions, and 4 range parameters, which leads to an action space of 756. 

\para{Reward function} The reward function reflects the quality of an executed action, while the agent is expected to be inspired by the reward to generate simple but expressive shape abstractions. The primary goal is to encourage the consistency between the generated primitive-based representation $\mathop{\cup}\limits_{i} P_i$ and the target shape $O$. We measure the consistency by the following two terms based on the intersection over union (IoU):
\begin{equation}
    \mathcal{I}_1 = IoU(\mathop{\cup}\limits_{i} P_i,O)   \quad\quad  \mathcal{I}_2 = \frac{1}{\mathcal{K}} \sum\limits_{P_i\in{\overline{\mathcal{P}}}} IoU(P_i,O),
\end{equation}
where $\mathcal{I}_1$ is the global IoU term and $\mathcal{I}_2$ is the local IoU term to encourage the agent to make each primitive cover more valid parts of the target shape; $\overline{\mathcal{P}}$ ($|\overline{\mathcal{P}}|=\mathcal{K}$) is the set of primitives that are not deleted yet. To favor simplicity, i.e., small number of primitives, we introduce a parsimony reward measured by the number of deleted primitives denoted by $\mathcal{N}$. Therefore, the reward function at the $k^{th}$ step is defined as

\begin{equation}
    R_k = (\mathcal{I}^k_1-\mathcal{I}^{k-1}_1)+ \alpha_1(\mathcal{I}^k_2 - \mathcal{I}^{k-1}_2)+ \alpha_2( \mathcal{N}^k-\mathcal{N}^{k-1}),
\label{eq:reward}
\end{equation}
where $\alpha_1$ and $\alpha_2$ are the weights to balance the last two terms. We set $R_k=-1$ once all the primitives are removed by the agent at $k^{th}$ step. The designed reward function motivates the agent to achieve higher volume coverage using larger and fewer primitives.

\subsection{Mesh Editing by Edge Loops}
\label{sec:mesh_agent}
An edge loop is a series of connected edges on the surface of an object that runs completely around the object and ends up at the starting point. It is an effective tool that plays a vital role in modeling software \cite{edgeloop2000}. Using edge loops, modelers can jointly edit a group of vertices and control an integral geometric unit instead of editing each vertex separately, which preserves the mesh regularity and improves the efficiency. Therefore, we make the Mesh-Agent learn mesh editing based on edge loops to produce higher mesh quality. 

\begin{figure}[!htb] 
    \centering
  \begin{overpic}[width=\linewidth]{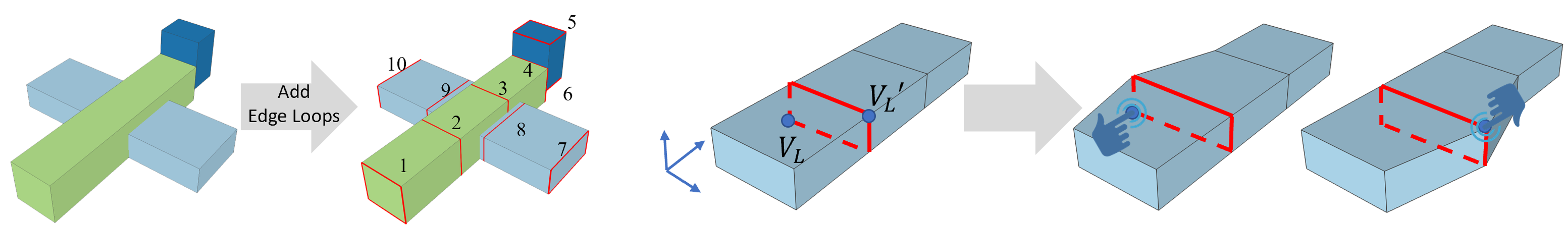}
        \put(165, -4) {Loop $L_i$}
        \put(225, -4) {Edit corner $V_L$}
        \put(290, -4) {Edit corner $V_L'$}
        \put(60, -13){(a)}
        \put(240, -13){(b)}
    \end{overpic}
  \caption{(a) We assign edge loops to the output primitives of the Prim-Agent for further mesh editing. Here, we show an example of adding $n=10$ edge loops to 3 primitives. (b) Two types of actions to operate an edge loop.}
  \label{fig:actions2}
\end{figure}
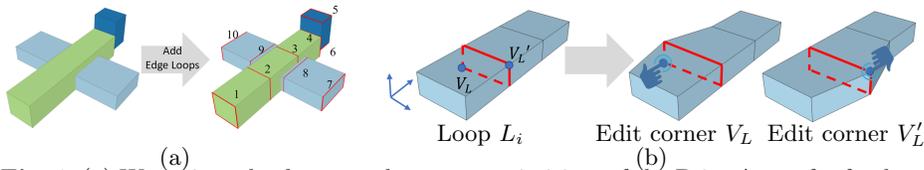

\para{Edge loop assignment} The output primitives from the last step do not have any edge loops. Thus we need to define edge loops on these primitives. For a primitive $P_i$, we choose the axis in which the longest cuboid side (principle direction) lies to assign the loop, while the loop planes are vertical to the chosen axis. We assign $n$ loops to $\mathcal{K}$ (not removed) cuboids; the number of loops assigned to a cuboid is proportional to its volume, while a larger cuboid will be assigned more loops. Each cuboid is assigned at least two loops on the boundaries. An example of edge loop assignment is shown in Fig. \ref{fig:actions2} (a). 

\para{State} We define the state by: (1) the input shape reference; (2) the updated edge loops at each iteration; (3) the step number represented by one-hot encoding. An edge loop $L_i$ is a rectangle defined by a six-tuple $(x_l,y_l,z_l,x'_l,y'_l,z'_l)$ which specifies its two diagonal corner points $V_L=(x_l,y_l,z_l)$, $V'_L=(x'_l,y'_l,z'_l)$.

\para{Action} As shown in Fig. \ref{fig:actions2} (b), we define two types of actions to operate a loop $L_i$: (1) drag the corner $V_L$; (2) drag the corner $V'_L$. For each type of action, we use six parameters $-3,-2,-1,1,2,3$ to control the range of movement on three axis directions. The number of edge loops we use is $n=10$ in this paper. In total, there are 10 edge loops, 2 types of actions, 3 moving directions and 6 range parameters, which leads to an action space of 360. 

\para{Reward function} The goal of this step is to encourage visual similarity of the edited mesh with the target shape, which can be measured by IoU. Accordingly, the reward is defined by the increments of IoU after executing an action.

\subsection{Virtual Expert}
\label{sec:virtualexpert}

Given such a huge action space, complex environment and long operation steps in this task, it is extremely difficult to train the modeling agents from scratch. However, collecting large scale sequence demonstration data from real experts can be expensive and the data are far from covering most scenarios. To address this problem, we propose an efficient heuristic algorithm as a virtual expert to generate the demonstration data. Note that the proposed algorithm is not for producing perfect actions used as ground-truth, but it can help the agents start the exploration with relatively better performance. More importantly, the agents are able to learn even better policies than imitating the virtual expert by the self-exploration in the RL phase (see the evaluation in Sec.~\ref{sec:discussion}).

For the primitive-based shape abstraction, we design an algorithm that outputs the actions as the following heuristics. We iteratively visit each primitive, test all the potential actions for the primitive and execute the one which can obtain the best reward. During the first half of the process, we do not consider any delete operations but only adjust the corners. This is to encourage all the primitives to fit the target shape first. Then in the second half, we allow deleting the primitives to eliminate redundancy. 

Similarly, for the edge loop editing, we iteratively visit each edge loop, test all the potential actions for the edge loop, and execute the one which can obtain the best reward.

\subsection{Agent Training Algorithm}
\label{sec:training_strategy}

Although using IL to warm up RL training has been researched in robotics \cite{dqfd2018}, directly applying off-the-shelf methods to train the agents for this problem domain does not produce good performance (see the experiments in Sec. \ref{sec:discussion}). Our task has the following unique challenges: (1) compared to the robotics tasks \cite{brockman2016openai} of which action space is usually less than 20, our agents need to handle over 1000 actions in a long sequence for modeling a 3D shape. This requires that the data from both ``expert'' and self-exploration should be organized and exploited effectively by the experience replay buffer. (2) The modeling demonstrations are generated by heuristics which are imperfect and monotonous, and thus the training scheme should not only use the ``expert'' to its fullest potential, but also enable the agents to escape from local optimum.

\begin{figure}[t]
    \centering
  \begin{overpic}[width=0.9\linewidth]{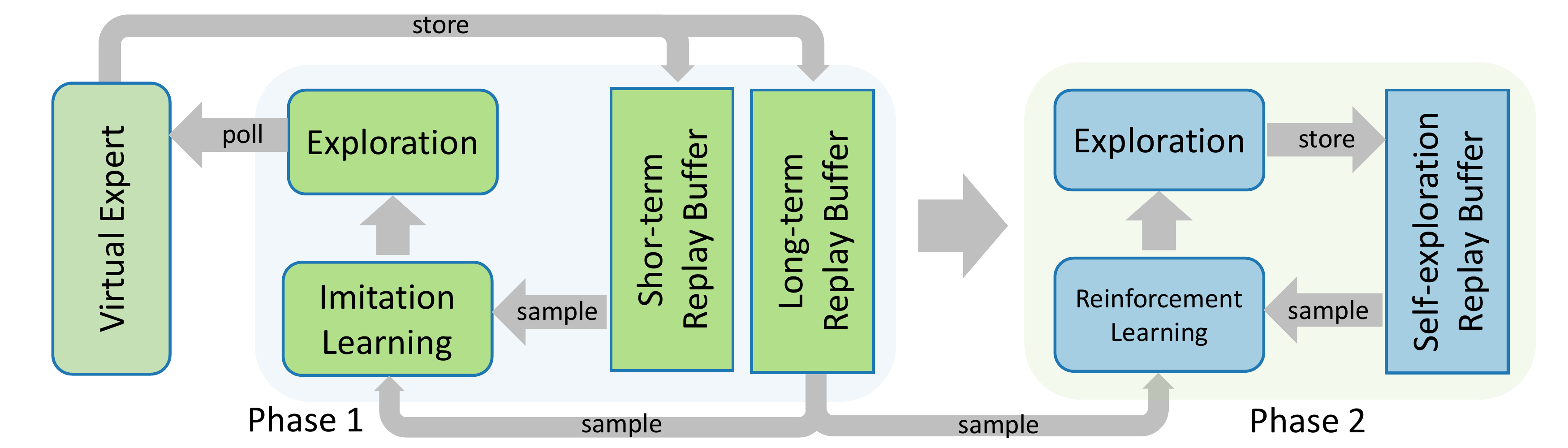}
    \end{overpic}
  \caption{Illustration of the architecture and the data flow of our training scheme. }
  \label{fig:train_phase}
\end{figure}

Therefore, in this section, we introduce a variant algorithm to train the modeling agents. The architecture of our training scheme is illustrated in Fig. \ref{fig:train_phase}. 

\para{Basic network} The basic network is based on the Double DQN (DDQN) \cite{van2016doubledqn} to predict the Q-values of the potential actions. The network outputs a set of action values $Q(s,\cdot;\theta)$ for an input state $s$, where $\theta$ are the parameters of the network. DDQN uses two separate Q-value estimators, i.e., current and target network, each of which is used to update the other. An experience is denoted as a tuple $\{s_k,a,R,s_{k+1}\}$ and the experiences will be stored in a replay buffer $\mathcal{D}$; the agent is trained by the data sampled from $\mathcal{D}$. The loss function for training DDQN is determined by temporal difference (TD) update:

\begin{equation}
    \mathcal{L}_{TD}(\theta) = ((R +\gamma Q(s_{k+1},a^{max}_{k+1};\theta' )-Q(s_k,a_k;\theta ))^2,\\
\end{equation}
where $R$ is the reward, $\gamma$ the discount factor, $a^{max}_{k+1}=argmax_a Q(s_{k+1},a;\theta)$, $\theta$ and $\theta'$ the parameters of current and target network respectively.

\para{Imitation learning by dataset aggregation} 
Our imitation learning process benefits from the idea of data aggregation (DAgger) \cite{ross2011dagger} which is an interactive guiding method.  A notable limitation of DAgger is that an expert has to be always available during training to provide additional feedback to the agent, making the training expensive and troublesome. However, benefiting from the developed virtual expert, we are able to guide the agent without additional cost by integrating the virtual expert into the training process. 

Different from the original DAgger, we use two replay buffers, named $\mathcal{D}^{demo}_{short}$ and $\mathcal{D}^{demo}_{long}$ for storing short-term and long-term experiences respectively. The $\mathcal{D}^{demo}_{short}$ only stores the experiences at the current iteration and will be emptied once an iteration is completed, while the $\mathcal{D}^{demo}_{long}$ stores all the accumulated experiences. At iteration $k$, we train a policy $\pi_k$ that mimics the ``expert'' on these demonstrations by equally sampling from both $\mathcal{D}^{demo}_{short}$ and $\mathcal{D}^{demo}_{long}$. Then we use the policy $\pi_{k}$ to generate new demonstrations, but re-label the actions using the heuristics of the virtual expert described in Sec. \ref{sec:virtualexpert}. 

Incorporating the virtual expert into DAgger, we poll the ``expert'' policy outside its original state space to make it iteratively produce new policies. Using double replay buffers provides a trade-off between learning and reviewing in the long sequence of decisions for shape modeling. The algorithm is detailed in Algorithm \ref{alg:dagger} with pseudo-code.

\begin{algorithm}[!htb]
Use the virtual expert algorithm to generate demonstrations $\mathcal{D}_0=\{(s_1,a_1),...,(s_M,a_M)\}$. \quad\\
Initialize $\mathcal{D}^{demo}_{short} \leftarrow \mathcal{D}_0$, $\mathcal{D}^{demo}_{long} \leftarrow \mathcal{D}_0$.\quad\\
Initialize $\pi_1$. \quad\\
\For{$k=1$ \text{to} $N$}
{
\quad\\
Train policy $\pi_{k}$ by equally sampling on both $\mathcal{D}^{demo}_{short}$ and $\mathcal{D}^{demo}_{long}$. \quad\\
Get dataset $\mathcal{D}_k=\{(s_1'),(s_2'),...,(s_M')\}$ by $\pi_{k}$. \quad\\
Label $\mathcal{D}_k$ with the actions given by the virtual expert algorithm. \quad\\
Empty short-term memory $\mathcal{D}^{demo}_{short} \leftarrow \varnothing$. \quad\\
Aggregate dataset $\mathcal{D}^{demo}_{long} \leftarrow \mathcal{D}^{demo}_{long} \cup \mathcal{D}_k$, $\mathcal{D}^{demo}_{short} \leftarrow \mathcal{D}_k$
}

\caption{DAgger with virtual expert using double replay buffers}
\label{alg:dagger}
\end{algorithm}

Similar to \cite{dqfd2018}, we apply a supervised loss to force the Q-value of the actions of ``expert'' to be higher than the other actions by at least a margin:
\begin{equation}
    \mathcal{L}_{S}(\theta)=\max \limits_{a\in A}(Q(s,a;\theta)+l(s,a_E,a)) - Q(s,a_E;\theta),
    \label{eq:superverised_loss}
\end{equation}
where $a_E$ is the action taken by the ``expert'' in state $s$ and $l(s,a_E,a)$ is a margin function that is a positive number when $a \ne a_E$ and is 0 when $a=a_E$. The final loss function used to update the network in the imitation learning phase is defined by jointly applying TD-loss and supervised loss:
\begin{equation}
    \mathcal{L}(\theta)=\mathcal{L}_{TD}(\theta)+\lambda \mathcal{L}_{S}(\theta).
    \label{eq:loss_function}
\end{equation}

\para{Reinforcement learning by self-exploration} Once the imitation learning is completed, the agents will have learned a reasonable initial policy.  Nevertheless, the heuristics of the virtual expert suffer from the local minimum and the demonstrations cannot cover all the situations the agents will encounter in the real system. Therefore, we make the agents interact with the environment and learn from their own experiences in a reinforcement paradigm. In this phase, we create a separate experience replay butter $\mathcal{D}^{self}$ to store only self-generated data during the exploration, and maintain the demonstration data in $\mathcal{D}^{demo}_{long}$. In each mini-batch, similar to the last step, we equally sample the experiences from $\mathcal{D}^{self}$ and $\mathcal{D}^{demo}_{long}$, and update the network only using TD-loss $\mathcal{L}_{TD}$. In this way, the agents retain a part of the memory from the ``expert'' but also gain new experiences by their own exploration. This allows the agents to potentially compare the actions learned from the ``expert'' and explored by themselves, and then make better decisions based on the accumulated reward in the practical environment.

\section{Experiments}
\subsection{Implementation Details}
\para{Network architecture} As shown in Fig.~\ref{fig:pipline}, for the Prim-Agent, the encoder is composed of three parallel streams: three 2D convolutional layers for the shape reference, two fully-connected (FC) layers followed by ReLU non-linearities for the primitive parameters, and one FC layer with ReLU for the step indicator. The three streams are concatenated and input to three FC layers with ReLU non-linearities for the first two layers, and the final layer outputs the Q-values for all actions. The Mesh-Agent adopts a similar architecture. The Prim-Agent is unrolled for 300 steps to operate the primitives and the Mesh-Agent 100 steps, while we have observed that more steps do not result in further improvement.

\para{Agent training} We first train the Prim-Agent and then use its output to train the Mesh-Agent. To learn a relatively consistent mapping from the modeling actions to the edge loops, we sort the edge loops into a canonical order. Each network is first trained by imitation and then by a reinforcement paradigm. The capacities of the replay buffer $\mathcal{D}^{demo}_{long}$ and $\mathcal{D}^{self}$ are 200,000 and 100,000 respectively, while the agents will over-write the old data in the buffers when they are full. Two agent networks are trained with batch size 64 and learning rate $8e^{-5}$. In the IL process, we perform DAgger for 4 iterations for each shape and the network is updated with 4000 mini-batches in each DAgger iteration. In the RL, we use $\epsilon=0.02$ for $\epsilon$-greedy exploration, $\tau=4000$ for the frequency at which to update the target network, and $\gamma=0.9$ for the discount factor. 

We use $\alpha_1=0.1$ and $\alpha_2=0.01$ to balance the terms in the reward function Eq.~\ref{eq:reward}, and $\lambda=1.0$ in the loss function Eq.~\ref{eq:loss_function}. The expert margin $l(s,a_E,a)$ in Eq.~\ref{eq:superverised_loss} is set to 0.8 when $a \ne a_E$. We observe sometimes the agents are stuck at a state and output repetitive actions; therefore, at each step, we force the agents to edit a different object, i.e., editing the $i^{th}$ ($i\in \{1,2,...,m\}$) primitive or loop at the $k^{th}$ step, where  $i=k\bmod m$. Also, the output of the Prim-Agent may have redundant or small primitives contained in the large ones, while we merge them to make the results cleaner and simpler. 

\subsection{Experimental Results}
Following the works for part-based representation of 3D shapes \cite{tulsiani2017learningboxprimitive, paschalidou2019superquadrics, sun2019abstraction}, we train our modeling agents on three shape categories separately. We collect a set of 3D shapes from ShapeNet \cite{chang2015shapenet}, i.e., airplane(600), guitar(600) and car(600), to train our network. We render a 128*128 depth map for each shape to serve as the reference. We use $10\%$ shapes from each category to generate the demonstrations for imitation learning.  To show the exploration as well as the generalization ability, in each category, we select 100 shapes that are either without demonstrations(50) or unseen(50) for testing. 

\begin{figure}[!htb] 
    \centering
  \begin{overpic}[width=0.88\linewidth]{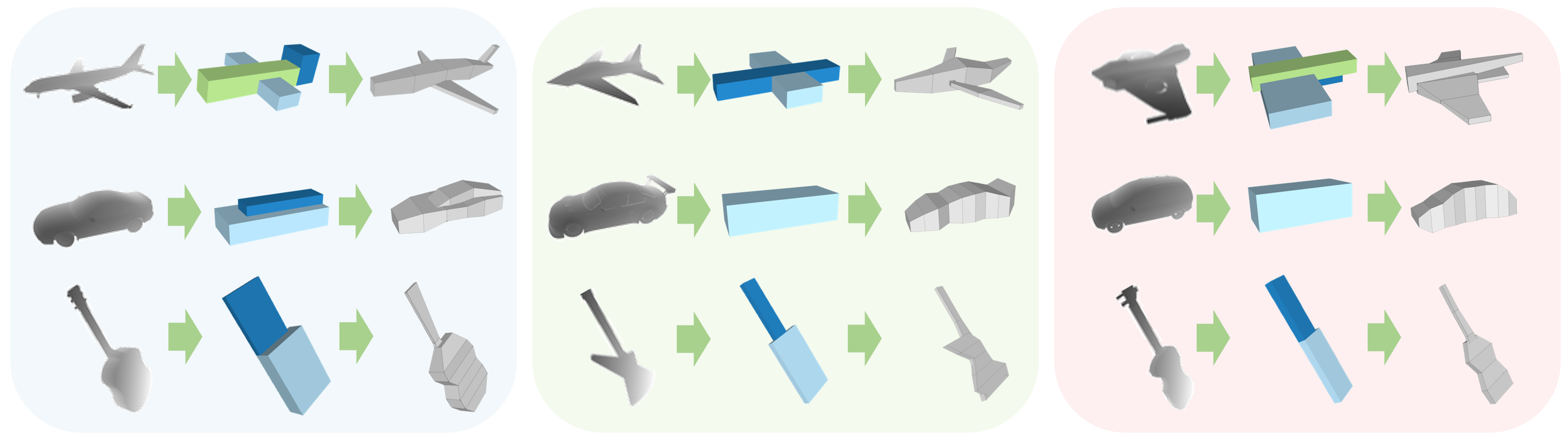}
    \end{overpic}
  \caption{Qualitative results of Prim-Agent and Mesh-Agent. Given a shape reference, the Prim-Agent first approximates the target shape using primitives and then the Mesh-Agent edits the meshes to create detailed geometry.}
  \label{fig:overall_results}
\end{figure}

\begin{figure}[!htb] 
\centering
  \begin{overpic}[width=0.88\linewidth]{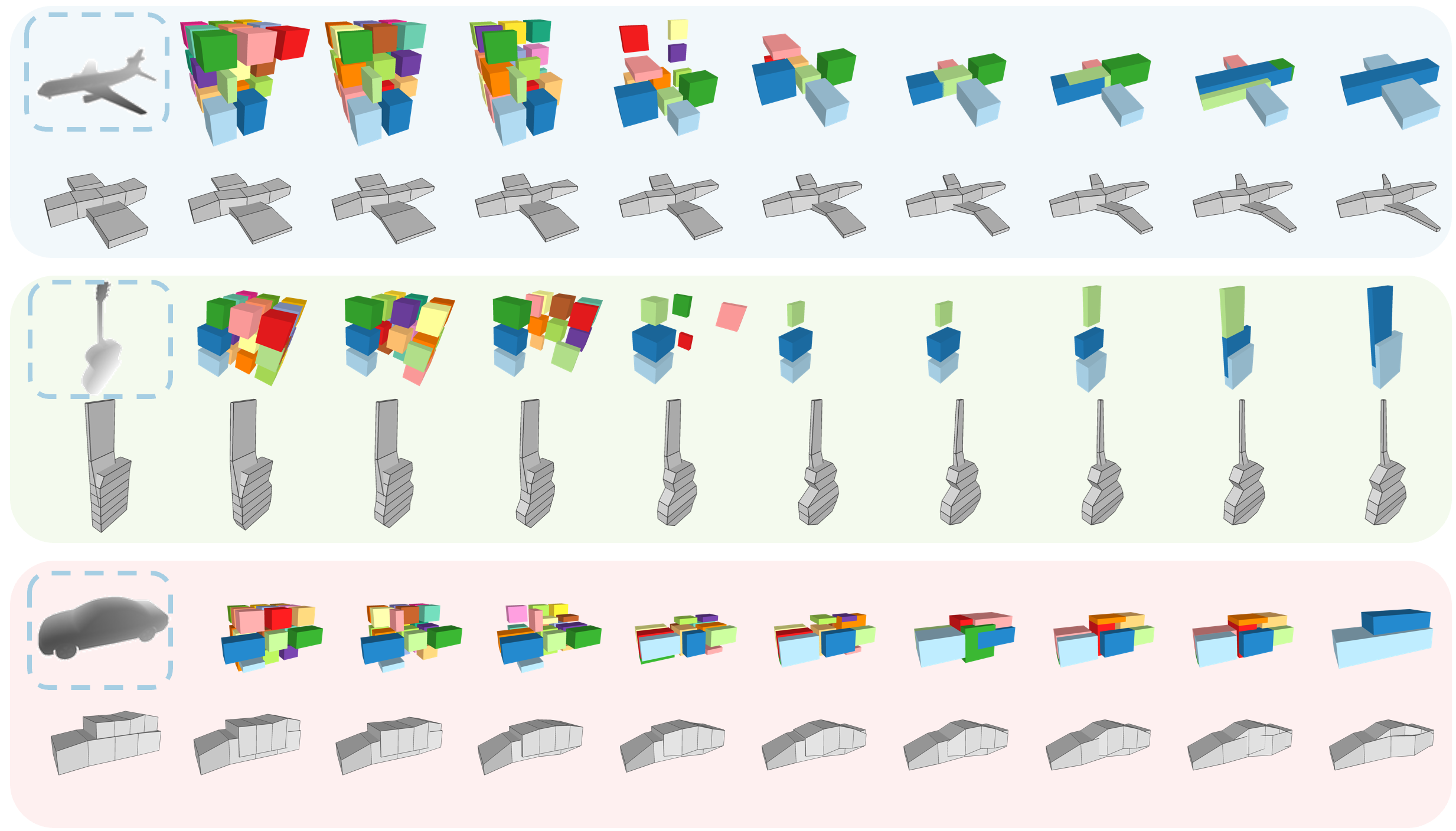}
    \end{overpic}
  \caption{The step-by-step procedure of 3D shape modeling. The first row of each sub-figure shows how the Prim-Agent approximates the target shape by operating the primitives (step 5, 10, 20, 40, 60, 80, 100, 200, 300). The second row shows the process of mesh editing by the Mesh-Agent (step 10, 20, 30, 40, 50, 60, 70, 80, 90, 100).}
  \label{fig:procedure_results}
\end{figure} 

We show a set of qualitative results in Fig.~\ref{fig:overall_results}. Given a depth map as shape reference, the Prim-Agent first approximates the target shape using primitives; then the Mesh-Agent takes as input the primitives and edits the meshes of the primitives to produce more detailed geometry. The procedure of the agents' modeling operation is visualized in Fig.~\ref{fig:procedure_results}. The agents show the power in understanding the part-based structure and capturing the fundamental geometry of the target shapes, and they are able to express such understanding by taking a sequence of interpretable actions. Also, the part-aware regular mesh can provide human modelers a reasonable initialization for further editing. 

\subsection{Discussions}
\label{sec:discussion}
\begin{figure}[!htb] 
  \begin{minipage}[c]{0.55\linewidth} 
    \begin{overpic}[width=\linewidth]{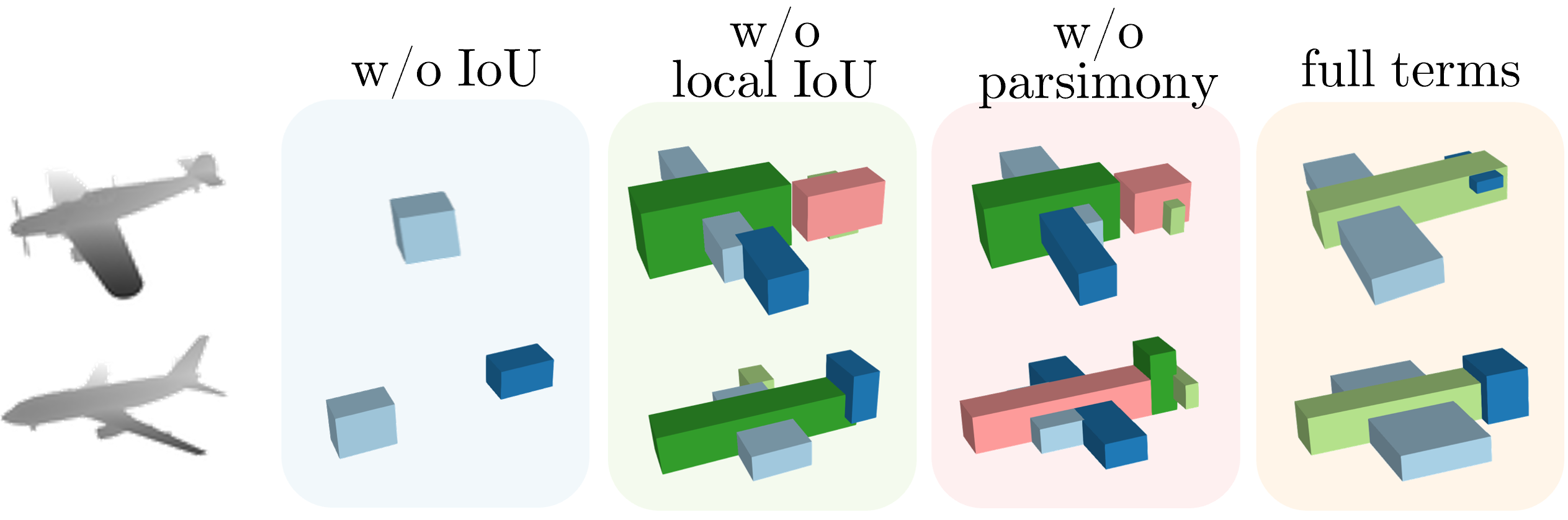}
    \put(100,-8){(a)}
    \put(250,-8){(b)}
    \end{overpic}
  \end{minipage}%
  \begin{minipage}[c]{0.4\linewidth}
\renewcommand{\arraystretch}{1.0}
    \begin{tabular}{c|c|c}\hline
    Config         & IoU        &  Prim Number     \\\hline
    w/o IoU        & 0.014      &      1.21        \\
    w/o local IoU  & 0.351      &      5.85        \\
    w/o parsimony  & 0.373     &      6.62        \\
    full terms     & 0.333      &      2.06         \\\hline
    \end{tabular}
  \end{minipage} 
\caption{Ablation study for the three terms in the reward function of the Prim-Agent. (a)~Qualitative results of using different configurations of the terms in the reward function. (b)~Quantitative evaluation; we show the average IoU and the numbers of the output primitives given different configurations.} 
\label{fig:ablation}
\end{figure}

\para{Reward function} Reward function is a key component for RL framework design. There are three terms in the reward function Eq.~\ref{eq:reward} for the Prim-Agent. To demonstrate the necessity of each term, we conduct an ablation study by alternatively removing each one and evaluating the performance of the agent. Fig.~\ref{fig:ablation}~(a) shows the qualitative results for different configurations. We also quantitatively report the average IoU and the average amount of the output primitives in Fig.~\ref{fig:ablation}~(b). Both qualitative and quantitative results show that using full terms is a trade-off between accuracy and parsimony, which can produce accurate but structurally simple representations that are more in line with human intuition.

\begin{figure}[!htb]
    \centering
  \begin{overpic}[width=\linewidth]{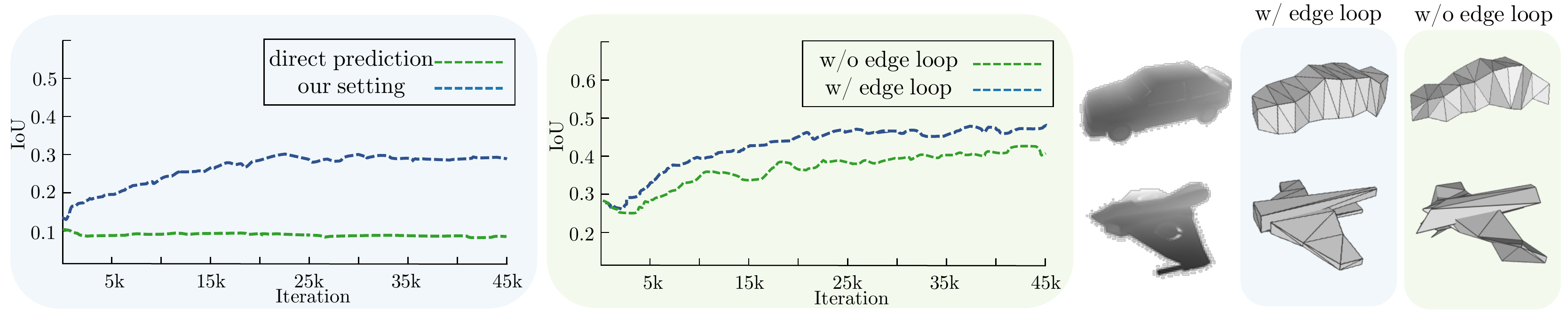}
    \put(60,-7){(a)}
    \put(175,-7){(b)}
    \put(287,-7){(c)}
    \end{overpic}
  \caption{Evaluations on the environment setting for Prim-Agent and Mesh-Agent. (a)~IoU over the course of training the Prim-Agent in different environment settings. (b)~IoU over the course of training the Mesh-Agent with and without using edge loops. (c)~Qualitative results produced by the Mesh-Agent with and without using edge loops; we show the triangulated meshes of the generated wireframes. }
  \label{fig:env_eval}
\end{figure}

\para{Does the Prim-Agent benefit from our environment?} We set up an environment where the Prim-Agent tweaks a set of pre-defined primitives to approximate the target shape step-by-step. A more straightforward way, however, is to make the agent directly predict the parameters of each primitive in a sequence. We evaluate the effect of these two environment settings on the agent for understanding the primitive-based structure. As shown in Fig. \ref{fig:env_eval} (a), the agent is unable to learn reasonable policies by directly predicting the primitives. The reason behind this is, in such an environment, the effective attempts are too sparse during exploration and the agent cannot be rewarded very often. Instead, in our environment setting, the task is decomposed into small action steps that the agent can simply do. The agent obtains gradual feedback and can be aware that the policy is getting better and closer. Therefore, the learning is progressive and smooth, which is advantageous to incentivize the agent to achieve the goal. 

\para{Do the edge loops help?} We use edge loops as the tool for geometry editing. To evaluate the advantages of our environment setting for the Mesh-Agent, we train a variant of the Mesh-Agent without using edge loops, where the agent edits each vertex separately. This leads to a doubled action space and uncorrelated operations between vertices. As shown in Fig.~\ref{fig:env_eval} (b) and (c), the agent using edge loops yields a lower modeling error and better mesh quality.

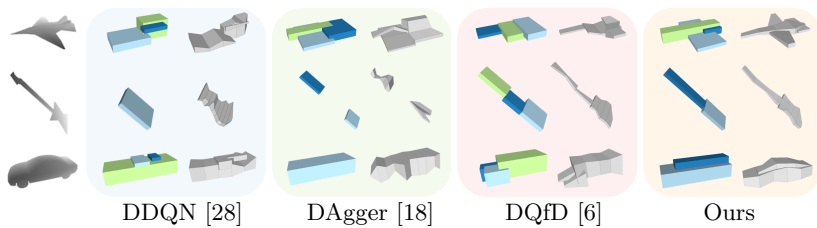
\begin{figure}[!htb] 
  \centering
  \begin{overpic}[width=0.9\linewidth]{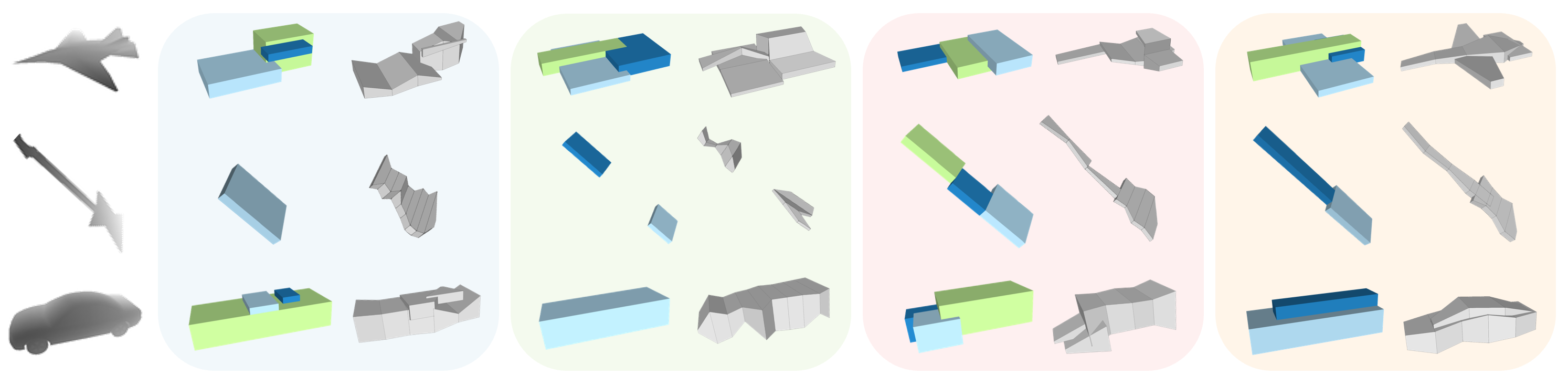}
  \put(45,-8){DDQN \cite{van2016doubledqn}}
  \put(115,-8){DAgger \cite{ross2011dagger}}
  \put(190,-8){DQfD \cite{dqfd2018}}
  \put(265, -8){Ours}
  \end{overpic}
  \caption{Qualitative comparison with related RL algorithms on the 3D modeling task. Our method gives better results, i.e., more structurally meaningful primitive-based representations and more regular and accurate meshes. }
  \label{fig:visual_cmp}
\end{figure} 

\begin{table}[!htb]
\centering
\scalebox{0.85}{
\begin{tabular}{c|ccc|ccc} \hline
                &\multicolumn{3}{c|}{Prim-Agent} & \multicolumn{3}{c}{Mesh-Agent} \\\hline
                                 & Airplane       & Guitar         & Car            & Airplane       & Guitar         & Car            \\\hline
DDQN (only RL)                   & 0.377          & 0.214          & 0.703          & -0.013         & -0.025         & 0.002          \\
DAgger (only interactive IL)     & 0.574          & 0.802          & 0.755          & 0.046          & 0.089          & 0.059          \\
DQfD (non-interactive IL + RL)   & 0.685          & 0.723          & 0.789          & 0.019          & 0.042          & 0.055          \\
DAgger* (double replay buffers)  & 0.725          & 0.954          & 0.897          & 0.048          & 0.105          & 0.065          \\
Ours (interactive IL + RL)       & \textbf{0.764} & \textbf{0.987} & \textbf{0.956} & \textbf{0.134} & \textbf{0.204} & \textbf{0.134}  \\\hline
\end{tabular}
}
\caption{Comparison with related learning algorithms. We report the average accumulated rewards gained by the agents on each category.}
\label{tab:cmp_rl}
\end{table}

\begin{table}[!htb]
\centering
\resizebox{\columnwidth}{!}{
\begin{tabular}{c|c|c|c|c|c|c|c|c|c|c|c|c}
\hline
&\multicolumn{6}{c|}{Prim-Agent} & \multicolumn{6}{c}{Mesh-Agent} \\\hline
&\multicolumn{2}{c|}{Airplane} &\multicolumn{2}{c|}{Guitar} &\multicolumn{2}{c|}{Car} &\multicolumn{2}{c|}{Airplane} &\multicolumn{2}{c|}{Guitar}&\multicolumn{2}{c}{Car} \\\hline
        & IoU            & CD              & IoU            & CD              & IoU            & CD              & IoU            & CD              & IoU            & CD              & IoU            & CD              \\\hline
DDQN    & 0.082          & 0.1165          & 0.094          & 0.1010          & 0.382          & 0.0812          & 0.069          & 0.1177          & 0.069          & 0.1092          & 0.384          & 0.0864          \\
DAgger  & 0.133          & 0.1068          & 0.202          & 0.0890          & 0.406          & 0.0761          & 0.179          & 0.0926          & 0.291          & 0.0804          & 0.466          & 0.0763          \\
DQfD    & 0.132          & 0.1112          & 0.196          & 0.0937          & 0.415          & 0.0749          & 0.151          & 0.1047          & 0.238          & 0.0796          & 0.471          & 0.0729          \\
DAgger* & 0.131          & 0.1104          & 0.275          & 0.0808          & 0.449          & 0.0778          & 0.179          & 0.0986          & 0.381          & 0.0598          & 0.514          & 0.0670          \\
Ours    & \textbf{0.179} & \textbf{0.0966} & \textbf{0.308} & \textbf{0.0595} & \textbf{0.481} & \textbf{0.0669} & \textbf{0.313} & \textbf{0.0917} & \textbf{0.512} & \textbf{0.0476} & \textbf{0.614} & \textbf{0.0532}
\\\hline
\end{tabular}
}
\caption{Quantitative evaluation on the shape reconstruction quality using additional metrics: IoU and Chamfer distance (CD).}
\label{tab:cmp_rl_shape_quality}
\end{table}

\para{Is our learning algorithm better than the others for 3D modeling?} In Sec. \ref{sec:training_strategy}, we introduce an algorithm to train the agents by combining heuristics, interactive IL and RL. Here, we provide an evaluation of the proposed learning algorithm with a comparison to using different related learning schemes. Table \ref{tab:cmp_rl} shows the average accumulated rewards across categories of different algorithms: (1) using the basic setting of DDQN \cite{van2016doubledqn} without an IL phase; (2) using the original DAgger \cite{ross2011dagger} algorithm with only supervised loss without an RL phase; (3) using DQfD algorithm \cite{dqfd2018}, which also combines IL and RL but the agent learns on fixed demonstrations rather than being interactively guided; (4) only using our improved DAgger with double replay buffers; (5) our training strategy described in Sec. \ref{sec:training_strategy}. Tables \ref{tab:cmp_rl_shape_quality} shows the evaluation on the shape reconstruction quality measured by the Chamfer distance (CD) and IoU. Also, we show the qualitative comparison results with these algorithms in Fig.~\ref{fig:visual_cmp}.  

Based on the qualitative and quantitative experiments, we can arrive at the following conclusions: (1) introducing simple heuristics of the virtual expert by IL significantly improves the performance, since the results show the modeling quality is unacceptable only using RL; (2) the final policy of our agents outperform the policy learned from the ``expert'', since our method obtains higher rewards than only imitating the ``expert''; (3) our learning approach can learn better polices and produce higher-quality modeling results than other algorithms.

\begin{figure}[!htb] 
    \centering
  \begin{overpic}[width=0.85\linewidth]{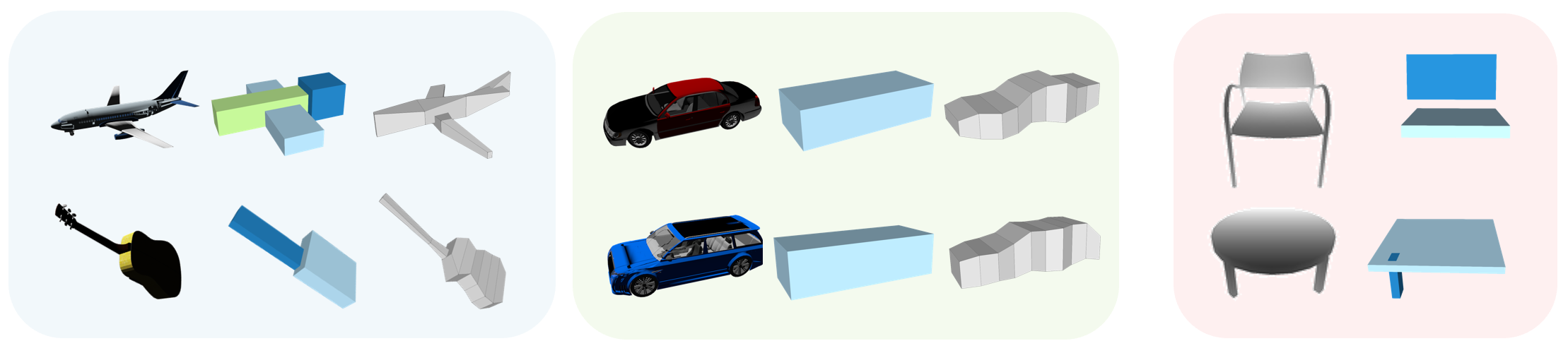}
    \put(100,-8){(a)}
    \put(250,-8){(b)}
    \end{overpic}
  \caption{(a) Modeling results using RGB images as reference. (b) Failure cases.}
  \label{fig:rgb-limitation}
\end{figure} 

\para{Can the agents work with other shape references?} We train the agents on a different type of reference, i.e., RGB images, without any modification. The average accumulated rewards obtained on different categories are 0.721, 0.877, 0.991 (Prim-Agent) and 0.120, 0.197, 0.135 (Mesh-Agent), which are similar to using depth maps. We also give some qualitative results in Fig. \ref{fig:rgb-limitation} (a).

\para{Limitations} A limitation of our method is, it fails to capture very detailed parts and thin structures of shapes. Fig. \ref{fig:rgb-limitation} (b) shows the results on a chair and a table model. Since the reward is too small when exploring the thin parts, the agent tends to neglect these parts to favor parsimony. A potential solution could be to develop a reward shaping scheme to increase the rewards at the thin parts.

\section{Conclusion}
In this work, we explore how to enable machines to model 3D shapes like human modelers using deep reinforcement learning. Mimicking the behavior of 3D artists, we propose a two-step RL framework, named Prim-Agent and Mesh-Agent respectively. Given a shape reference, the Prim-Agent first parses the target shape into a primitive-based representation, and then the Mesh-Agent edits the meshes of the primitives to create fundamental geometry. To effectively train the modeling agents, we introduce an algorithm that jointly combines heuristic policy, IL and RL. The experiments demonstrate that the proposed RL framework is able to learn good policies for modeling 3D shapes.

Overall, we believe that our method is an important first stepping stone towards learning modeling actions in artist-based 3D modeling. Ultimately, we hope to achieve conditional and purely generative agents that cover various modeling operations, which can be integrated into modeling software as an assistant to guide real modelers, such as giving step-wise suggestions for beginners or interacting with modelers to edit the shape cooperatively, thus significantly reducing content creation cost, for instance in games, movies, or AR/VR settings.

\para{Acknowledgements} We thank Roy Subhayan and Agrawal Dhruv for their help on data preprocessing and Angela Dai for the voice-over of the video. We also thank Armen Avetisyan,  Changjian Li, Nenglun Chen, Zhiming Cui for their discussions and comments.  This work was supported by a TUM-IAS Rudolf M\"o{\ss}bauer Fellowship, the ERC Starting Grant \textit{Scan2CAD} (804724), and the German Research Foundation (DFG) Grant \textit{Making Machine Learning on Static and Dynamic 3D Data Practical}.

\clearpage
%
%

\bibliographystyle{splncs04}
\bibliography{reference}

\title{Modeling 3D Shapes by Reinforcement Learning Supplementary Material} 


\titlerunning{Modeling 3D Shapes by Reinforcement Learning Supplementary Material} 

%
\author{Cheng Lin\inst{1,2}\and
Tingxiang Fan\inst{1}\and
Wenping Wang\inst{1}\and
Matthias Nie\ss ner\inst{2}}
\authorrunning{C. Lin et al.}

%

\institute{The University of Hong Kong\and Technical University of Munich 
 \\
}
\maketitle

\section{Network Architecture}
\begin{figure}[!htb] 
    \centering
  \begin{overpic}[width=0.8\linewidth]{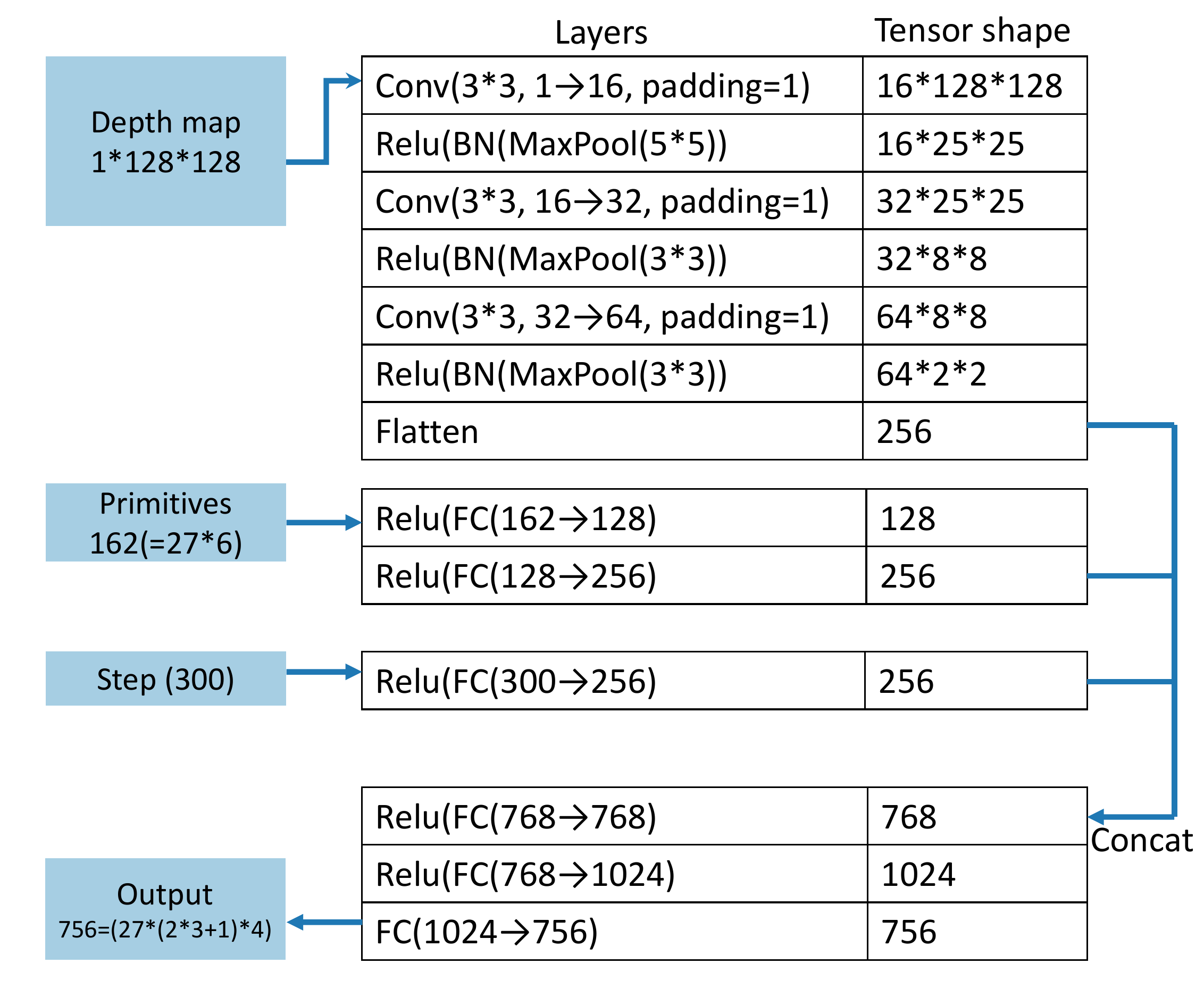}
    \end{overpic}
  \caption{The detailed network architecture of the Prim-Agent. BN: Batch Normalization Layer; FC: Fully Connected Layer.}
  \label{fig:prim_net}
\end{figure}

\begin{figure}[!htb] 
    \centering
  \begin{overpic}[width=0.8\linewidth]{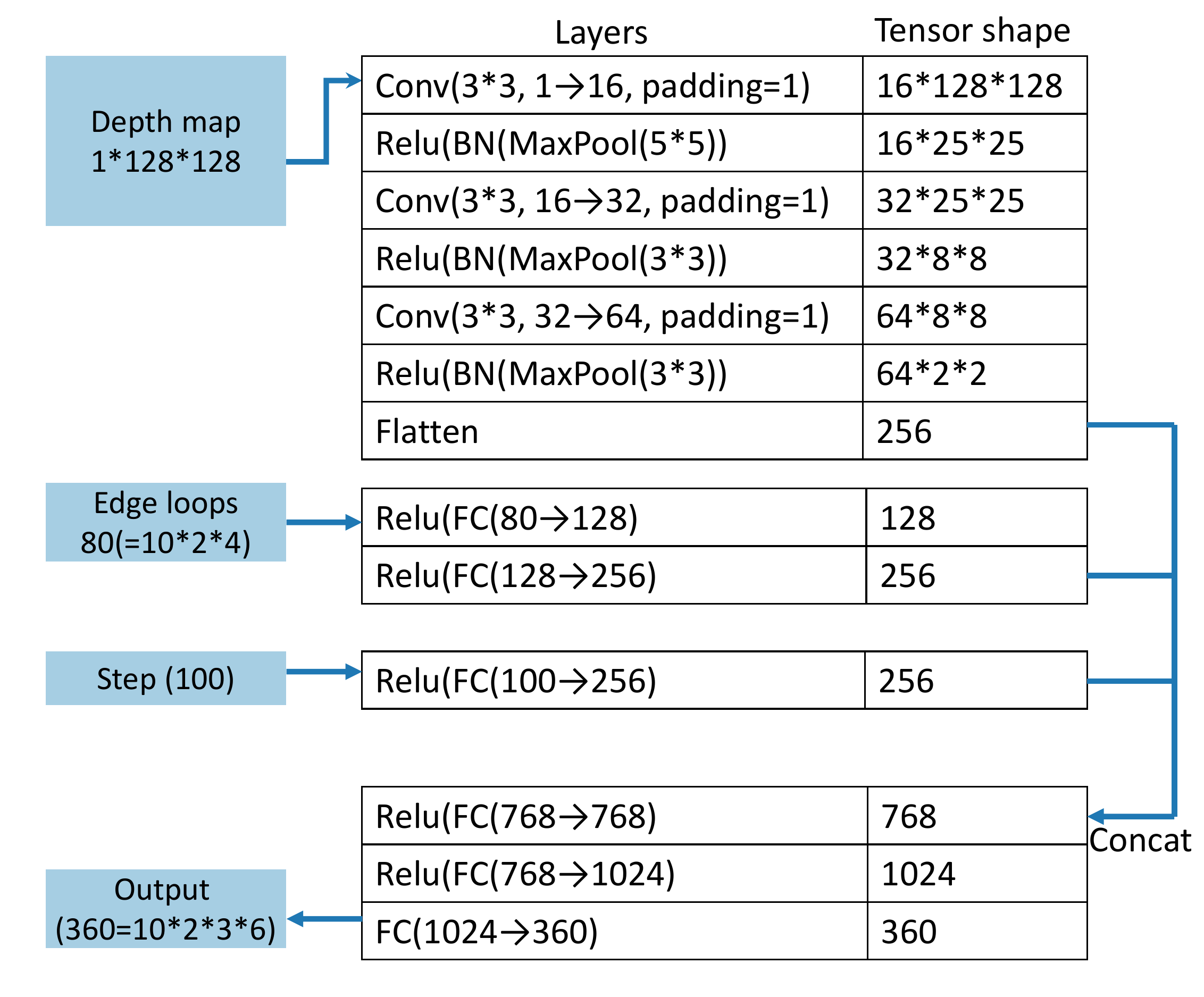}
    \end{overpic}
  \caption{The detailed network architecture of the Mesh-Agent. BN: Batch Normalization Layer; FC: Fully Connected Layer. The feature dimension of a loop point is 4, i.e., $(x_l, y_l, z_l, a)$ where $a\in \{0,1,2\}$ additionally indicates the axis the loop plane is vertical to.}
  \label{fig:mesh_net}
\end{figure}

Fig. \ref{fig:prim_net} and Fig. \ref{fig:mesh_net} show the detailed architecture of the Prim-Agent and the Mesh-Agent respectively. We also indicate the shape of the tensor output from each layer.


\section{Illustration of the Choices of Method Design}
\subsection{Solution Space Reduction}
We should note that it is not trivial for an RL agent to learn to model 3D shapes. The biggest challenge is that the action space has enormous modeling operations, and many of them are irrelevant. In the paper, there are in total 1116 different actions and the network will be unrolled for 400 steps, which leads to a huge solution space of $1116^{400}$. Therefore, the exploration to find good policies will be extremely difficult. Here we summarize the key ideas to make this task feasible.

\para{Divide the solution space} Inspired by the hierarchical understanding of human modelers, we divide these operations into two categories, i.e., primitive-based operations and mesh-based operations, to reinforce more connections between different actions. Therefore, we propose two sperate agents, i.e., Prim-Agent and Mesh-Agent. The solution space is split down into $756^{300}$ and $360^{100}$ respectively for each step and the difficulty of learning is reduced as well.

\para{Learn an initial policy} As described in the paper, the agents are first trained to imitate the demonstrations generated by heuristics. Second, with the learned initial policy, the agents then learn in an RL paradigm by collecting the rewards. Since most of the actions in the huge solution space produce very poor performance which is meaningless, the initial policy can significantly reduce the number of exploration of poor performance.

\para{Restrict the actions in each step} The strategies mentioned above can already help the agents learn reasonable policies, but the training efficiency is still fairly low. Also, we observe sometimes the agents are stuck at a state and output repetitive actions; therefore, at each step, we force the agents to edit a different primitive or loop from the last step.

To overcome these two issues, the strategy we adopt is that, at the $k^{th}$ step, we force the agents to only choose the actions that can operate the $i^{th}$ primitive (or loop), where  $i=k\bmod m$ and $m$ is the number of the primitives (or loops). The action space is further narrowed down in each step, and the agents will not be stuck at a repetitive action.

\subsection{Local IoU Reward}
The local IoU reward encourages the Prim-Agent to make each primitive cover more valid parts of a target shape, which will make the primitives overlap first. Therefore, deleting an overlapped primitive will gain high sparsity reward without losing much accuracy. Without the local IoU reward, since simplicity conflicts with accuracy, the agents cannot be motivated to balance the parsimony and the accuracy to give structurally meaningful and simple representations.

\subsection{Double Replay Buffers for IL}
If we only use one buffer, the expert’s new demonstrations are mixed together with the old ones. This may lead to inadequate learning of the new experiences, given that the old and new data are sampled together but the old ones are sufficiently learned in previous iterations. Therefore, we propose to use two buffers: the short-term replay buffer $\mathcal{D}^{demo}_{short}$ is for learning the newest demonstrations, while the long-term one $\mathcal{D}^{demo}_{long}$ is for reviewing the histories. This is shown to be more effective.  

\section{Virtual Expert Algorithm}
We give the detailed algorithm of the virtual expert for the Prim-Agent in Algorithm~\ref{alg:virtual_expert} with pseudo-code. We iteratively visit each primitive, test all the potential actions for the primitive and execute the one which can obtain the best reward. Note the selection of actions is divided into two stages: (1) during the first half of the process, we do not consider any delete operations but only edit the corners; (2) in the second half, deleting a primitive is allowed. 

For the Mesh-Agent, we iteratively visit each edge loop, test all the potential actions, and execute the one which can obtain the best reward. Note there is only one stage for the ``expert'' of mesh editing. 

\begin{algorithm}[!htb]
\KwIn{ $m$ cuboid primitives $ \mathcal{P}=\{P_1,P_2,...,P_m\}$; target shape $O$; maximal step $N_{max}$ }
\KwOut{a sequence of actions $\mathcal{A}=\{a_1,a_2,...,a_N\}$}
\Repeat{$Step=N_{max}$}{

        \For{each $P_i\in P$}
        {   
            $Step$++
            
            \If{$Step$ $\le 0.5*N_{max}$}
            {
            find the action $a$ which has the highest reward to tweak a cuboid corner\quad\\
            }
            \Else
            {
            find the action $a$ which has the highest reward to tweak a cuboid corner or delete a cuboid\quad\\
            }
            
            execute and output the action $a$
            
            update the state $s$
            
        }
      }
\caption{Virtual Expert for Primitive-based Shape Abstraction}
\label{alg:virtual_expert}
\end{algorithm}

\section{Primitive Merging}
\label{sec:merging}
Even though we have introduced a parsimony term in the reward function, the output of the Prim-Agent may still have some small or redundant primitives. We design a simple algorithm to merge these primitives as follows.

We define a graph $G$ for the output primitives. In this graph, each node represents a primitive $P_i$. The merging of $P_i(V_i,V_i')$ and $P_j(V_j,V_j')$ will lead to a new primitive $P_{ij}(\min\{V_i,V_j\}, \max
\{V_i',V_j'\})$. Two nodes $P_i$ and $P_j$ will be connected by an edge if $IoU(P_i \bigcup P_j, P_{ij}) \ge \epsilon$.

We compute the connected components for the graph $G$ and then merge all the primitives in the same connected components into a single primitive. The merging process is performed for two iterations, while $\epsilon$ is set to 0.85 and 0.90 respectively in each iteration.

\section{Edge Loop Assignment}

Given $M'$ primitives and $N$ edge loops, we assign the edge loops onto the longest axis of each primitive while the loops are uniformly distributed in that direction. The number of loops $E(P_k)$ assigned to a primitive $P_k$ is determined by 
\begin{equation}
    E(P_k)= \max\{\lceil{N\frac{V(P_k)}{\sum\limits_{i} {V(P_i)}}+0.5}\rceil, 2\} ,
\end{equation}
where $V(P_i)$ is the volume for the primitive $P_i$ and $i\in\{1,2,...,M'\}$.  It can be seen that the number of loops assigned to a cuboid is proportional to its volume; thus a larger cuboid will be assigned more loops. Each cuboid is assigned at least two loops on the boundaries. When dealing with the last primitive $P_{M'}$, we directly assign all the remaining unallocated loops on to $P_{M'}$.

\end{document}